# A Framework for Predicting Phishing Websites Using Neural Networks

A.Martin[1], Na.Ba.Anutthamaa[2], M.Sathyavathy[3], Marie Manjari Saint Francois[4], Dr.Prasanna Venkatesan[5]

[1] Banking Technology, Pondicherry University
Puducherry, India

[2] Information Technology, Sri Manakula Vinayagar Engineering College
Puducherry, India

[3] Information Technology, Sri Manakula Vinayagar Engineering College
Puducherry, India

[4] Information Technology, Sri Manakula Vinayagar Engineering College
Puducherry, India

[5] Banking Technology, Pondicherry University
Puducherry, India

## Abstract

In India many people are now dependent on online banking. This raises security concerns as the banking websites are forged and fraud can be committed by identity theft. These forged websites are called as Phishing websites and created by malicious people to mimic web pages of real websites and it attempts to defraud people of their personal information. Detecting and identifying phishing websites is a really complex and dynamic problem involving many factors and criteria. This paper discusses about the prediction of phishing websites using neural networks. A neural network is a multilayer system which reduces the error and increases the performance. This paper describes a framework to better classify and predict the phishing sites using neural networks.

***Key Words:*** *Phishing, Neural Networks, Classification, Learning, Phishing Detection*.

## 1. Introduction

Cybercrime refers to any crime that involves a computer and a network, where the computers may or may not have played an instrumental part in the commission of a crime. Computer crime encompasses a broad range of potentially illegal activities. Generally, however, it may be divided into one of two types of categories: (1) crimes that target computer networks or devices directly and (2) crimes facilitated by computer networks or devices, the primary target of which is independent of the computer network or device. Examples for cybercrimes are fraud, spam, cyber terrorism and phishing.

Phishing is a type of online fraud in which a scam artist uses an e-mail or website to illicitly obtain confidential information. It is a semantic attack which targets the user rather than the computer. It is a relatively new internet crime. The phishing problem is a hard problem because of the fact that it is very easy for an attacker to create an exact replica of a good banking website which looks very convincing to users. The communication (usually email) directs the user to visit a web site where they are asked to update personal information, such as passwords and credit card, social security, and bank account numbers that the legitimate organization already has .

## 2. Phishing Scams

There are many ways in which someone can use phishing to social engineer someone. For example, someone can manipulate a website address to make it look like you are going to a legitimate website, when in fact you are going to a website hosted by a criminal.

The process of phishing involves five steps namely, planning, setup, attack, collection and identity theft and fraud. During the planning stage the phishers decide which business to target and determine how to get e-mail addresses for the customers of that business. They often use the same mass-mailing and address collection techniques as spammers. In the setup stage after they know which business to spoof and who their victims are, the phishers create methods for delivering the message and collecting the data. Most often, this involves e-mail addresses and a web page. The attack stage is the step people are most familiar with - the phisher sends a phony message that appears to be from a reputable source. The collection stage is the one in which phishers record the information entered by victims into Web pages or popup windows. The final stage is the Identity theft and Fraud where the phishers use the information they've gathered to make illegal purchases or otherwise commit fraud. As many as a fourth of the victims never fully recover.

If the phisher wants to coordinate another attack, he evaluates the successes and failures of the completed scam and begins the cycle again. Phishing scams take advantages of software and security weaknesses on both the client and server sides .

## 3. Literature Review

### 3.1 Approaches to prevent phishing

Despite growing efforts to educate users and create better detection tools, users are still very susceptible to phishing attacks. Unfortunately, due to the nature of the attacks, it is very difficult to estimate the number of people who actually fall victim. A report by Gartner estimated the costs at $1,244 per victim, an increase over the $257 they cited in a 2004 report [9]. In 2007, Moore and Clayton estimated the number of phishing victims by examining web server logs. They estimated that 311,449 people fall for phishing scams annually, costing around 350 million dollars. There are several promising defending approaches to this problem reported earlier.

The first approach is [11] to stop phishing at the email level, since most current phishing attacks use broadcast email (spam) to lure victims to a phishing website .Another approach [10] is to use security toolbars. The phishing filter in IE8 is a toolbar approach with more features such as blocking the user's activity with a detected phishing site.

A third approach is to visually differentiate the phishing sites from the spoofed legitimate sites. Dynamic Security Skins [12] proposes to use a randomly generated visual hash to customize the browser window or web form elements to indicate the successfully authenticated sites.

A fourth approach is two-factor authentication, which ensures that the user not only knows a secret but also presents a security token. However, this approach is a server-side solution. Sensitive information that is not related to a specific site, *e.g.*, credit card information and SSN (Social Security Number), cannot be protected by this approach either. Many industrial antiphishing products use toolbars in Web browsers, but some researchers have shown that security tool bars don't effectively prevent phishing attacks.

The Passpet system, created by Yee et al. in 2006, uses indicators so that users know they are at a previously trusted website. Since all of these proposals require the use of complicated third-party tools, its unclear how many users will actually benefit from them. The newest version of Microsoft's Internet Explorer supports Extended Validation (EV) certificates, coloring the URL bar green and displaying the name of the company. However, a recent study found that EV certificates did not make users less fall for phishing attacks.

### 3.2 Cryptographic Identity Verification Method

Proposals have been made for a scheme that utilises a cryptographic identity verification method that lets remote Web servers prove their identities. However, this proposal requires changes to the entire Web infrastructure (both servers and clients), so it can succeed only if the entire industry supports it. The crypto module encrypts and decrypts data received from the memory under the control of the central processing unit. The security service station likewise also contains a central processing unit including a memory, and a crypto module. This station also contains a comparator for comparing personal identification information with reference personal identification information. Both kinds of information are transmitted to the station from the terminal.

## 4. Case Studies on E-Banking Phishing Sites

Phishing occurs despite the growing efforts that are taken to educate users and users are still very much susceptible to phishing attacks. Even the display of Extended Validation certification did not decrease the percentage of users who are made victims of phishing attacks. To stress on the need to predict phishing two case studies have been conducted.

### 4.1 Website Phishing

Consider the original website and the phished website of a bank namely, the State Bank of India (SBI) which is involved in e-banking. Unless the user is a known visitor of the site it is not possible for him/her to identify the authentication of the site based on its look and feel.When we take a close look at the two sites some differences can be observed, (1) URL is different - The URL of the original site is **www.onlinesbi.com** [4] and the URL of the phished website is **www.sbionline.com** [5] and (2) Validation of the EV SSL certificate - Extended Validation Secure Sockets Layer (SSL) Certificates are special SSL Certificates that work with high security Web browsers to clearly identify a Web site's organizational identity.Extended Validation (EV) helps you make sure a Web site is genuine and verified. In original websites, the address bar turns green indicating that the site is secured by an EV certificate.

The following figure 1 is the original website of SBI,

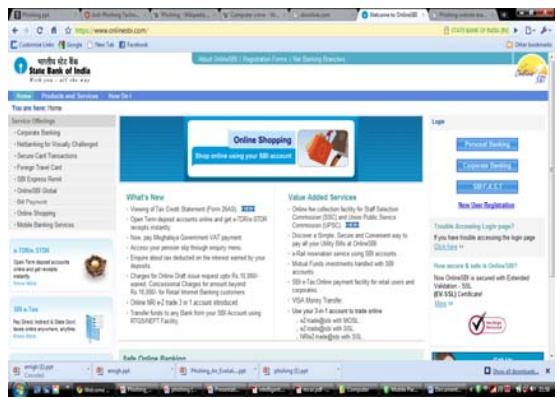

Figure:1 – Original website of SBI

The figure 2 shows the phished website of SBI ,

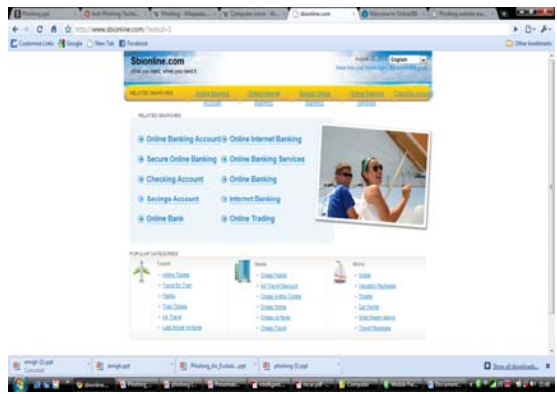

Figure:2 – Phished website of SBI

For example, using IP address instead of domain name, http instead of https, poor design, spelling errors, absence of SSL padlock icon and phony security certificate.

The deceiving email was sent to 120 employees informing them that their e-banking accounts are at the risk of being hacked and requested them to log into their account through fake link attached to our email using their usual customer ID and password to verify their balance and then log out normally.

The results of the above experiment are shown in the table 1.The results clearly indicate that target phishing factor is extremely dangerous since almost half of the employees who responded were victimized; particularly, trained employees such as those of IT Department and IT Auditors. Increasing the awareness of all users of e-banking regarding this risk factor is highly recommended.

TABLE : 1- PHISHING WEBSITE EXPERIMENT

| Response to Phishing Website | No. of Employees |
|---|---|
| Interacted positively (IT Department) | 8 |
| Interacted positively (Other Departments) | 44 |
| Interacted negatively (Incorrect info) | 28 |
| Interacted negatively (No response) | 40 |
| **Total** | 120 |

### 4.2 Phone phishing experiment

A group of 50 employees were contacted by female colleges assigned to lure them into giving away their personal e-banking accounts user name and password (through social and friendly phone conversation with a deceiving purpose in mind). The results were beyond expectations; many of the employees fell for the trick.After conducting friendly conversation with them for some time, our team managed to seduce them into giving away their internet banking credentials for fake reasons. Some of these lame reasons included checking their privileges and accessibility, or for checking its integrity and connectivity with the web server for maintenance purposes, account security and privacy assurance…etc.

TABLE : 2- PHONE PHISHING EXPERIMENT

| Response to phone phishing | No. of employees |
|---|---|
| Giving away their full ebanking credentials(user name & Password) | 16 |
| Giving away only their ebanking user name without password | 8 |
| Refused to reveal their credentials | 26 |
| **Total** | 50 |

To assure the authenticity of our request and to give it a social dimensional trend, our team had to contact them repeatedly for about three or four time. As shown in above table, we managed to deceive 16 out of the 50 employees to give away their full e-banking credentials which represented 32% of the sample.

A total of 16% (8 employees) agreed to give their user name only and refrained from giving away their Passwords under any circumstances or excuses what so ever. The remaining 52% (26 employees) were very cautious and declined to reveal any information regarding their credentials over the phone. An overview of the results reveals the high risk of social engineering security factor. Social engineering constitutes a direct internal threat to e-banking web services since its hacks directly into the accounts of e-bank customers. The results also show the direct need to increase the awareness of customers not to fall victims of this kind of threat that can lead to devastating results.

The result of the above experiment is shown in the table 2.The percentage in the below table is considered high since the experiment was conducted on the staff members of bank, who are supposed to be highly educated with regard to the risks of electronic banking services.

## 5. Phishing Characteristics and Indicators

Phishing in e-banking is prevalent nowadays. The phishing problem is a hard problem because of the fact that it is very easy for an attacker to create an exact replica of a good banking site, which looks very convincing to users.

Based on case studies conducted 27 features and indicators were gathered and clustered them into six criteria [1]. Those six criteria are URL & domain identity, Security & encryption, Source code & java script, Page style & contents, Web address bar and Social human factor.

TABLE : 3- PHISHING INDICATORS WITH ITS CRITERIA

| CRITERIA | N | PHISHING INDICATORS |
|---|---|---|
| URL & Domain Identity | 1 | Using IP address |
| | 2 | Abnormal request URL |
| | 3 | Abnormal URL of anchor |
| | 4 | Abnormal DNS record |
| | 5 | Abnormal URL |
| Security & Encryption | 1 | Using SSLcertificate |
| | 2 | Certificate authority |
| | 3 | Abnormal cookie |
| | 4 | Distinguished names certificate |
| Source code & java script | 1 | Redirect pages |
| | 2 | Straddling attack |
| | 3 | Pharming attack |
| | 4 | OnMouseOver to hide the link |
| | 5 | Server form handler |
| Page style & Contents | 1 | Spelling errors |
| | 2 | Copying website |
| | 3 | Using forms with Submit button |
| | 4 | Using pop-up windows |
| | 5 | Disabling right click |
| Web address bar | 1 | Long url address |
| | 2 | Replacing similar char for URL |
| | 3 | Adding a prefix or suffix |
| | 4 | Using the @ symbol to confuse |
| | 5 | Using the hexadecimal char codes |
| Social human factor | 1 | Emphasis on security |
| | 2 | Public generation salutation |
| | 3 | Buying time to access accounts |

## 6. Neural Networks and Phishing Prediction

We are going to utilize neural network techniques in our new e-banking phishing website detection model as shown in table 3 to find the most important phishing features and significant patterns of phishing characteristic or factors in the e-banking phishing website archive data. Each indicator will range between the input values genuine, doubtful and legitimate. Using these values rules will be formed and the network will be trained to give output that ranges between Very legitimate, legitimate, suspicious, phishy and very phishy.

An artificial neural network (ANN), usually called neural network (NN), is a mathematical model or computational model that is inspired by the structure and/or functional aspects of biological neural networks. A neural network consists of an interconnected group of artificial neurons, and it processes information using a connectionist approach to computation. In most cases an ANN is an adaptive system that changes its structure based on external or internal information that flows through the network during the learning phase. The most interesting feature in neural networks is the possibility of learning. The network learns when examples with known results are presented to it. The weighting factors are adjusted by an algorithm to bring the final output closer to known result.

Each unit in neural network performs a simple computation: it receives signals from its input links and computes a new activation level that it sends along each of its output links. The computation of the activation level is based on the values of each input signal received from a neighboring node, and the weights on each input link. The computation is split into two components. First is a *linear* component, called the input function, **in$_i$** that computes the weighted sum of the unit's input values. Second is a *nonlinear* component called the **activation function,** *g,* that transforms the weighted sum into the final value that serves as the unit's activation value, $a_i$. The total weighted input is the sum of the input activations times their respective weights:

$$in_i = \sum_j w_{j,i} a_j = w_i a_j$$

Most neural network learning algorithms, including the follow the current-best-hypothesis In this case, the hypothesis is a network, defined by the current values of the weights. The initial network has randomly assigned weights, usually from the range [-0.5, 0.5]. The network is then updated to try to make it consistent with the examples. This is done by making small adjustments in the weights to reduce the difference between the observed and predicted values. Typically, the updating process is divided into **epochs.** Each epoch involves updating all the weights for all the examples. For perceptrons, the weight update rule is particularly simple. If the predicted output for the single output unit is *O,* and the correct output should be *T,* then the error is given by

$$Err = T-O$$

If the error is positive, then we need to increase *O;* if it is negative, we need to decrease *0*. Now each input unit contributes $W_j a_i$ to the total input, so if $a_i$ is positive, an increase in $W_j$ will tend to increase *O*, and if $a_i$ is negative, an increase in $W_j a_i$ will tend to decrease *O*. Thus, we can achieve the effect we want with the following rule:

$$w_j \longleftarrow w_j + a \times a_j \times Err$$

where the term a is a constant called the **learning rate.**

In multilayer networks, there are many weights connecting each input to an output, and each of these weights contributes to more than one output. The back-propagation algorithm is a sensible approach to dividing the contribution of each weight. At the output layer, the weight update rule is given below. There are two differences: the activation of the hidden unit $a_j$ is used instead of the input value; and the rule contains a term for the gradient of the activation function. If Err$_i$ is the error *(T - O)* at the output node, then the weight update rule for the link from unity to unit *i* is

$$w_{j,i} \longleftarrow w_{j,i} + a \times a_j \times Err_i g'(in_i)$$

where *g'* is the derivative of the activation function *g*. We will find it convenient to define a new
error term A$_i$ which for output nodes is defined as A$_i$ = Err$_i$ *g'(in$_i$)* The update rule then becomes

$$w_{j,i} \longleftarrow w_{j,i} + a \times a_j \times A_i$$

For updating the connections between the input units and the hidden units, we need to define a quantity analogous to the error term for output nodes. Here is where we do the error backpropagation.

The hidden node *j* is "responsible" for some fraction of the error A, in each of the output nodes to which it connects. Thus, the A$_j$ values are divided according to the strength of the connection between the hidden node and the output node, and propagated back to provide the A, values for the hidden layer. The propagation rule for the A values is the following

$$A_{j,i} = g'(in_j) \sum w_{j,i} A_j$$

Now the weight update rule for the weights between the inputs and the hidden layer is almost
identical to the update rule for the output layer:

$$w_{k,j} \longleftarrow w_{k,j} + a \times I_k A_j$$

The below given algorithm explains how the weight update rule works for the output layer and hidden layer. The algorithms has *networks, examples* and *a* as inputs where *a* is the *learning rate*.

Function WEIGHT_UPDATE (network, examples, a)

returns a network with modified weights

Inputs: network, a multilayer network

      examples, a set of input/output pairs

      a, the learning rate

repeat

    for each e in examples do

$$O \leftarrow RUN\_NETWORK(network, I^e)$$

$$Err^e \leftarrow T^e - O$$

$$W_{j,i} \leftarrow W_{j,i} + O \times a_j \times Err^e \times g'(in_i)$$

    for each subsequent layer in network do

$$A_j \leftarrow g'(in_j) \sum_i W_{j,i} A_i$$

$$W_{k,j} \leftarrow W_{k,j} + a \times I_k \times A_j$$

This algorithm explains for updating weights in a multi-layer network.

Initially all the phishing website details are collected and stored in the phishing website archive. Then it is sent to a preprocessor to convert into machine understandable format.

The result is then stored as records in the database. The database also stores configuration parameters (the 27 phishing indicators that are being extracted from the code). Using the data collected in the database, rules are generated to detect the website phishing rate using the neural network techniques.

Once the neural network has been created it needs to be trained with the existing data in the archive. One way of doing this is initialize the neural net with random weights and then feed it a series of inputs.

We then check to see what its output is and adjust the weights accordingly so that whenever it sees something looking like the existing data it outputs the same result as that data. This is shown in the figure 3,

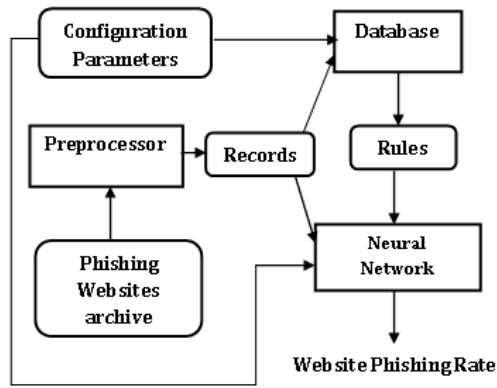

Figure:3- A model for detecting phishing websites

## 7. Website Phishing Training Data Sets

For our implementation we plan to use two publicly available datasets to test: the "phishtank" from the phishtank.com[8] .The PhishTank database records the URL for the suspected website that has been reported, the time of that report, and sometimes further detail such as the screenshots of the website, and is publicly available.

The Anti Phishing Working Group (APWG) which maintains a "Phishing Archive" describing phishing attacks. In addition, 27 features are used to train and test the classifiers [7]. We will use a series of short scripts to programmatically extract the above features, and store them in an excel sheet for quick reference. The age of the dataset is the most significant problem, which is particularly relevant with the phishing corpus. E-banking Phishing websites are short-lived, often lasting only in the order of 48 hours. Some of our features can therefore not be extracted from older websites, making our tests difficult. The average phishing site stays live for approximately 2.25 days.

## 8. Conclusion

The prediction of phishing websites is essential and this can be done using neural networks. For the prediction of phishing websites, earlier works were done using various data mining classification algorithms were used but the error rate of those algorithms were very high [2]. When an element of the neural networks fails, it can continue without any problem because of its parallel nature. Thus performance can be made better by considering neural networks as it reduces the error and gives better classification. We believe that this framework works better and gives a lower error rate.